\pdfoutput=1
\documentclass[10pt,letterpaper]{article}

\usepackage{cogsci}
\usepackage[margin=1in]{geometry} 
\usepackage{microtype}
\usepackage{graphicx}
\usepackage{subfigure}
\usepackage{booktabs} 
\usepackage{nicefrac}       
\usepackage{microtype}      
\usepackage{float}
\usepackage{hyperref}
\hypersetup{colorlinks,allcolors=purple}
\usepackage{algorithm}
\usepackage{amsmath}
\usepackage{graphicx}
\usepackage[table,xcdraw]{xcolor}
\usepackage{gensymb}
\usepackage{stmaryrd}
\usepackage{amssymb}
\usepackage{todonotes}
\usepackage{comment}

\cogscifinalcopy 

\usepackage{pslatex}
\usepackage{apacite}

\title{Identifying concept libraries from language about object structure}

 \author{
    {\large \bf Catherine Wong$^{\star}$}, {\large \bf William P. McCarthy$^{\star}$}$^{2}$, {\large \bf Gabriel Grand$^{\star}$}$^{1}$,
    {\large \bf Yoni Friedman}$^1$, \\
    {\large \bf Joshua B. Tenenbaum}$^{1}$, {\large \bf Jacob Andreas}$^{1}$, {\large \bf Robert D. Hawkins}$^{3}$ \and {\large \bf Judith E. Fan}$^2$ \\ 
    $^1$MIT,
    $^2$ University of California San Diego,
    $^3$Princeton Neuroscience Institute \\
    $^{\star}$denotes equal contribution, correspondence to \texttt{catwong@mit.edu}}

\begin{document}

\maketitle

\begin{abstract} 
Our understanding of the visual world goes beyond naming objects, encompassing our ability to parse objects into meaningful parts, attributes, and relations. 
In this work, we leverage natural language descriptions for a diverse set of 2K procedurally generated objects to identify the parts people use and the principles leading these parts to be favored over others.
We formalize our problem as search over a space of program libraries that contain different part concepts, using tools from machine translation to evaluate how well programs expressed in each library align to human language. By combining naturalistic language at scale with structured program representations, we discover a fundamental information-theoretic tradeoff governing the part concepts people name: people favor a lexicon that allows concise descriptions of each object, while also minimizing the size of the lexicon itself.

\end{abstract}

\noindent The world is filled with a great variety of objects, yet people have little difficulty making sense of them. 
Presented with a novel object, people can readily identify its parts \shortcite{schyns1994ontogeny}, guess its function \shortcite{tversky1984objects}, and refer to it unambiguously \shortcite{hawkins2020characterizing}. 
These abilities rest on the capacity to robustly connect features of the external world to a rich library of mental concepts describing not just whole objects, but their parts and how they are arranged \shortcite{miller2013language, landau1993and, rosch1975family,mukherjee2019communicating}. 

\begin{figure*}[ht!]
  \begin{center}
  \includegraphics[width=1.02\linewidth]{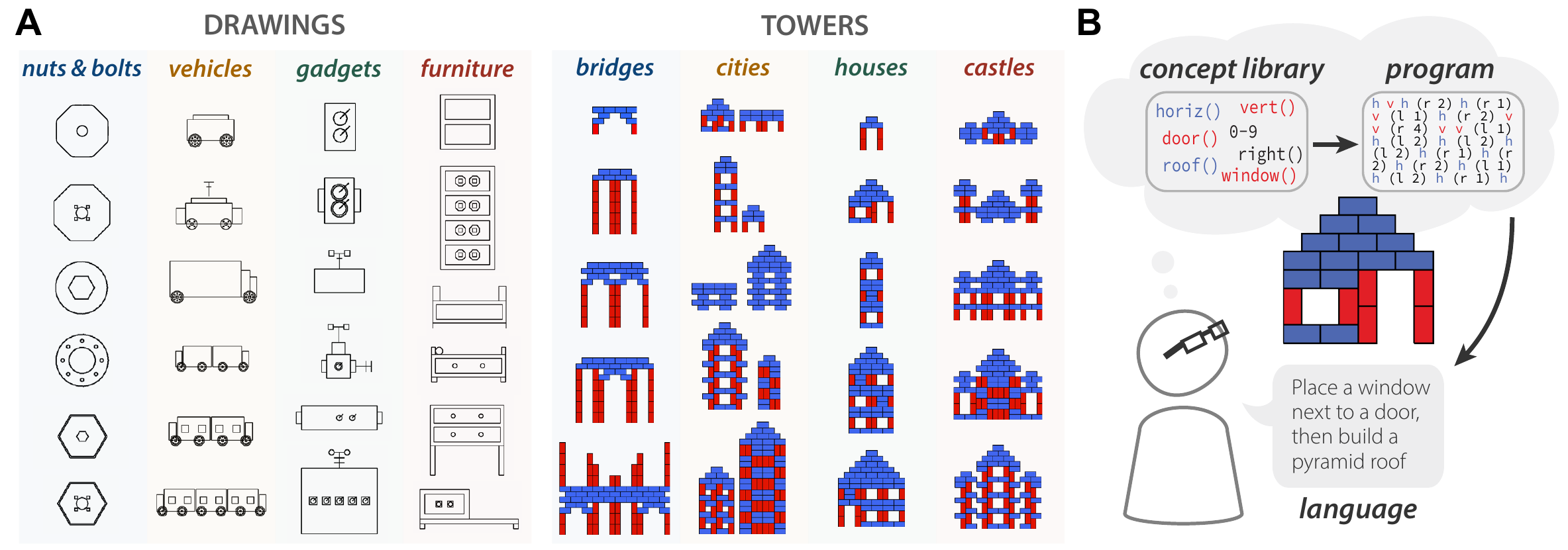}
  \caption{(A) Example objects from the \textit{Drawings} and \textit{Towers} domains. Each domain contains 4 subdomains of 250 novel objects. Each domain and subdomain was designed to include high variation over the type and number of base primitives (i.e., shapes, blocks). (B) This work aims to infer the latent concept library that people are using to decompose complex objects into parts, where objects are represented by executable graphics programs.}
  \label{fig:task}
  \end{center}
 \end{figure*}

For example, consider the bottom-most \emph{gadget} in Fig.~\ref{fig:task}A: even though this object does not correspond to a familiar category, we might say that it contains a row of \texttt{buttons} or \texttt{dials}, and that it is topped by an \texttt{antenna} or a \texttt{knob}.
But just as we do not have a pre-existing concept for every object we encounter, we do not have a concept corresponding to every part: in Fig.~\ref{fig:task}A, for example, most people do not have a concept corresponding to a \texttt{row of exactly five dials}. 
Indeed, a complex object can be decomposed in a huge number of different ways, but people are likely to favor only a tiny subset of them.
What characterizes the set of part concepts that people do use? Why these, and not others?

Identifying which parts people use to parse visual objects has been a core goal for classic theories of perceptual organization
\shortcite{palmer1977hierarchical,marr1978representation,hoffman1984parts,biederman1987recognition}
and continues to pose challenges for modern vision models \cite{mo2019partnet,bear2020learning}.
But how can we tell whether any of these proposals actually explain visual object understanding? 
Empirical tests of these theories have generally relied upon simple discrimination tasks rather than richer behavioral readouts \shortcite{tversky1989parts, markman1988children}, limiting their ability to evaluate correspondences between a candidate object representation and the full set of parts and relations that people can identify.

Natural language offers a powerful window into our conceptual representations, given abundant evidence that our vocabularies have been shaped to efficiently communicate about the concepts we find relevant \shortcite{regier201511,kirby2015compression,zaslavsky2018efficient,sun2021seeing}. In this work, our goal is to leverage \textit{naturalistic language production} to identify the conceptual libraries of parts and relations used for visual object understanding, using libraries of symbolic \textit{program} components to model how these concepts are mentally represented (Fig.~\ref{fig:task}B). In this framework, each library instantiates a different hypothesis about the underlying inventory of part concepts that people are using to decompose visual objects.  


In Part I, we describe our strategy for creating a diverse collection of novel objects generated using graphics programs (Fig. ~\ref{fig:task}A), and for eliciting open-ended descriptions of these objects. Analyzing these descriptions reveals hallmarks of these concept libraries: people produce longer descriptions to describe more complex objects, and invoke different part concepts to describe objects from different categories. 
In Part II, we refine this picture with a formal library identification model that measures the correspondence between language and candidate program libraries containing part concepts of varying complexity, building on recent work in program library discovery
\shortcite{ellis2020dreamcoder,tian2020learning, wang2021learning,wong2021leveraging}. This model reveals a deeper information-theoretic principle governing the part concepts people invoke in language: they reflect a fundamental trade-off between the complexity of a concept library and the complexity of objects represented using that library. 

\section{Part I: Eliciting language about \\ object structure} \label{sec-part-i}

Our central aim is to identify the library of part concepts that people invoke to decompose objects.
Towards this end, we needed a sufficiently large and varied collection of objects, and a naturalistic task for eliciting detailed descriptions of their structure. 


\subsection{Methods}

\paragraph{Participants}
We recruited 465 participants from Prolific to complete the task. 
Participants provided informed consent and were paid approximately \$15 per hour. 

\begin{figure*}[ht!]
  \begin{center}
  \includegraphics[width=0.99\linewidth]{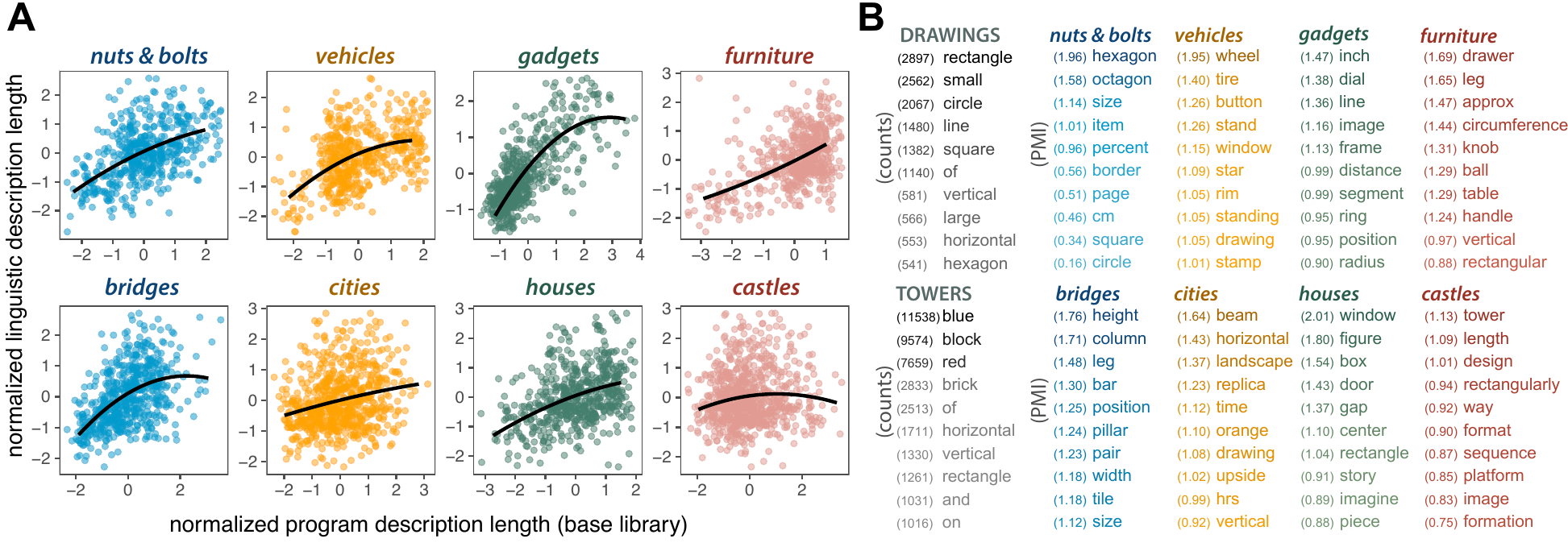}
  \caption{(A) Relationship between length of base-library programs and length of linguistic descriptions. (B) \textit{Left:} Top-10 words that appeared most frequently in descriptions for each domain. \textit{Right:} Top-10 words with highest pointwise mutual information (PMI) within each subdomain.}
  \label{fig:words}
  \end{center}
\end{figure*}

\paragraph{Stimuli}

To ensure that we had a sufficiently large and diverse collection of objects, we developed a hierarchical procedure for synthesizing complex configurations of shapes. 
Taking inspiration from recent work employing line drawings and block towers to investigate how people learn and represent the compositional structure of objects \shortcite{tian2020learning, mccarthy2021learning,wang2021learning}, we defined two stimulus \textit{domains}, distinguished by the set of base shape primitives used to generate them (Fig.~\ref{fig:task}A).
\textit{Drawings} are composed of simple geometric curves (i.e., \texttt{line}, \texttt{circle}) and are evocative of familiar object categories; \textit{Towers} are composed of rectangular blocks (i.e., horizontal and vertical dominoes) and are evocative of simple architectural models.

To investigate the degree to which people invoked category-specific part concepts to describe these objects, rather than the same set of ``atomic'' base primitives in all cases, we further defined four \textit{subdomains} nested within each domain. 
Within \textit{Drawings}, these were informally designated as \textit{nuts \& bolts}, \textit{vehicles}, \textit{gadgets}, and \textit{furniture}; and within \textit{Towers}, as \textit{bridges}, \textit{cities}, \textit{houses}, and \textit{castles} (Fig.~\ref{fig:task}A).
For each subdomain, we procedurally generated 250 unique examples, hierarchically composing the base primitives into increasingly complex, recursively defined parts. 
A \texttt{dresser}, for example, is composed of \texttt{drawers}, which are in turn composed of a \texttt{panel} and \texttt{knobs}, themselves defined by combining \texttt{circles} and \texttt{lines}.
In sum, this procedure yielded a varied collection of 2000 object stimuli: 1000 Drawings and 1000 Towers, each accompanied by a graphics program that can be used to regenerate it in terms of the base primitives.

\paragraph{Task procedure}

Each participant was instructed to provide step-by-step instructions for how to ``draw'' or ``build'' 10 different ``drawings'' or ``models'' sampled from a \textit{single} subdomain.
At the start of each session, participants were first familiarized with the general characteristics of the subdomain by viewing 25 examples (none of which then appeared during the main experiment, and none of which were accompanied by any linguistic labels for the subdomain).
Throughout the session, they were also shown the 7 upcoming objects they would be asked to describe, to provide  concurrent information about how objects varied within the subdomain.
Because we were primarily focused on interrogating which part descriptors people invoke, we designed the text-entry interface to encourage participants to describe each step by composing a \textit{what}-phrase and a \textit{where}-phrase, which were entered into separate text boxes. 
Participants could include as many instruction steps as they deemed necessary and there was no trial time limit.


\paragraph{Language preprocessing} 
To investigate the content of the instructions generated by participants, we used the spaCy NLP library to extract and lemmatize words, including part-of-speech (POS) tagging to remove determiners and punctuation. We also replaced common typos (``sqaure,'' ``cirlce,'' etc.) and spelling variations with their canonical spellings in US English.

\subsection{Results}

\paragraph{People use more words for more complex objects}
The simplest way that object structure may be exposed in language is through description complexity.
We consider three possibilities.
First, insofar as participants decompose all objects into the same number of parts, regardless of how complex these parts are, the length of their descriptions would be predicted to remain \emph{constant} over a wide range of objects.
Second, if participants tend to decompose objects into a set of commonly recurring parts, and mention each part, the length of their descriptions would be predicted to positively correlate with object complexity: the more parts, the longer the description. 
A third possibility is that there is a systematic but non-linear relationship between object complexity and linguistic description length \cite{sun2021seeing}, consistent with a compromise between the first two strategies. 

We operationalize object complexity here as the length of the (base) graphics program that generated it.
We measure the length of linguistic descriptions as the number of words provided in the \textit{what} phrases (ignoring for now spatial language in the \emph{where} phrases).
To tease apart the above hypotheses, we fit a mixed-effects model to predict linguistic description length from graphics program length, including random intercepts for participants and random effects of program length at the participant level.
We observed a significant main effect of program length ($t(318)=14.8, p < 0.001$ across all subdomains), providing strong evidence against the first view. 
We also found that a model including an additional quadratic effect of program length, allowing for a non-linear relationship, significantly improved the fit ($\chi^2(3)=38.6$), although the strength of this relationship varied across subdomains (Fig. \ref{fig:words}A).
These findings suggest that people generally use more words to describe more complex objects, but the strength and nature of this relationship can vary widely across object categories. 

\paragraph{People use different parts for different subdomains} 
If description length scales with object complexity (expressed in the base library), a natural possibility is that speakers are simply providing descriptions at the level of those low-level primitives.
For example, they may be giving block-by-block instructions for towers and line-by-line instructions for the drawings.
In this case, we would not expect differences in the distribution of words used across subdomains (e.g. ``bridges'' and ``houses'' would both be described in terms of the same red and blue blocks).
Alternatively, if speakers generate descriptions at higher levels of conceptual abstraction --- for example, in terms of ``pillars'' or ``windows'' --- we would expect their language to reflect the varying part structure of the subdomains.
To assess these competing hypotheses, we computed the pointwise mutual information (PMI) for each unique word $w$ in the language data with respect to the four subdomains $d$  (Fig.~\ref{fig:words}B):

\begin{equation} \label{eq:pmi}
PMI(w) = \log \dfrac{p(w, d)}{p(w)p(d)}
\end{equation}

Intuitively, PMI is high for words that occur more frequently in a particular subdomain (numerator) than would be expected given the overall prevalence of the word across subdomains and the amount of language data in each subdomain (denominator).
This analysis revealed highly specialized vocabularies used for particular subdomains, but not others (e.g., \textit{drawer} and \textit{knob} in the \textit{furniture} subdomain), suggesting that participants did invoke subdomain-specific part concepts to some extent.

To better evaluate whether these highly diagnostic words reflected more systematic differences in word usage across subdomains, we computed the Jensen-Shannon distance (JSD) between the word frequency distributions in each set of subdomains, aggregating across all trials in that subdomain.
This metric is zero when two distribution are identical and large when two distributions are far apart.
We compared the the mean of all pairwise JSDs to a null distribution generated by randomly permuting the subdomain group of each trial. 
We found that the distance between subdomains was significantly greater than  expected under the null (\textit{Drawings}: $d = 0.439$, $p < 0.001$; \textit{Towers}: $d = 0.328$, $p < 0.001$).
Taken together, these analyses indicate that people may choose distinct labels to describe visually similar parts depending on the rest of the scene (e.g. a circle may be a \emph{knob} in one domain and a \emph{wheel} in another domain), even when simple graphics primitives would have been sufficient.

\section{Part II: Identifying concepts from language}\label{sec-part-ii}
The results so far suggest that people invoke subdomain-specific part concepts when describing the objects in our stimulus set, such as \texttt{knobs} and \texttt{drawers}, or \texttt{windows} and \texttt{doors}.
What accounts for observed preferences for this lexicon --- how many and which part concepts do people have names for? 

In this section, we formalize this library identification problem by modeling the correspondence between people's vocabularies and a space of candidate concept libraries, each containing part concepts at varying levels of complexity. We describe a procedure for constructing candidate libraries based on the hierarchical structure of each subdomain. We then introduce a library-to-vocabulary alignment model that measures how well programs written in each library predict people's object descriptions \shortcite{wong2021leveraging}. 

Prior work suggests that people use language that efficiently compresses concepts into words \shortcite{regier201511,kirby2015compression,zaslavsky2018efficient,sun2021seeing}. Our model allows us to derive an information-theoretic account of lexical choice in our object descriptions, which formally links language to efficient communication of an object's underlying conceptual representation -- we find that people favor a lexicon that trades off between concise descriptions of objects on average, and the size of the overall concept libraries.





\subsection{Methods}
\paragraph{Modeling a space of candidate concept libraries} 

\begin{figure}[t]
  \begin{center}
  \includegraphics[width=0.99\linewidth]{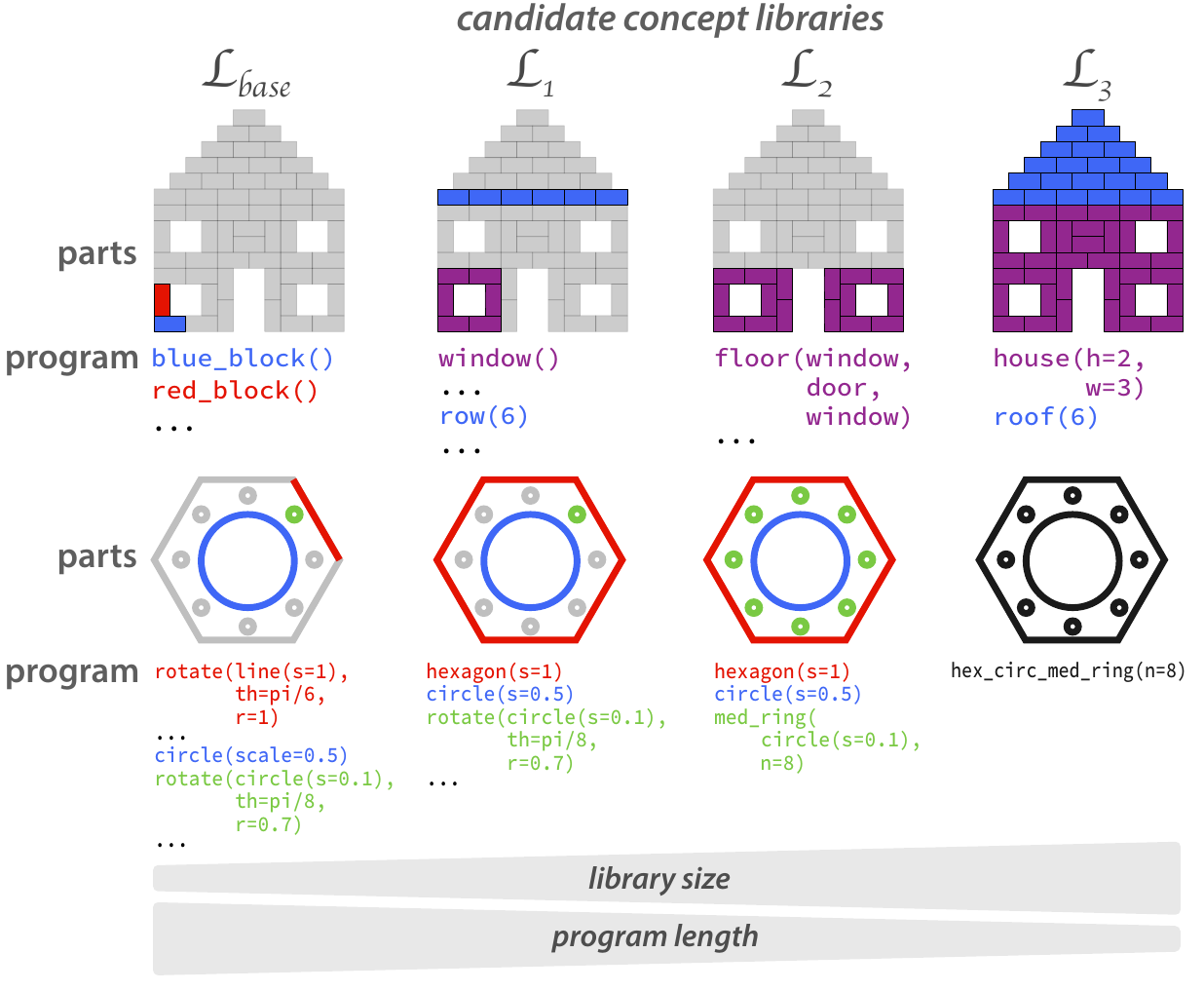} 
  \caption{Graphics libraries were defined by progressively adding subroutines at higher levels of abstraction, resulting in more efficient expression of any particular program at the expense of a larger library.}
  \label{fig:library_gallery}
  \end{center}
\end{figure}

By design, the objects in our stimulus set are highly structured, having been generated through the hierarchical combination of increasingly complex parts.
However, the corresponding graphics programs that recreate them were written using a concept library containing only the base primitives ($\mathcal{L}_{base}$): \texttt{blocks} and \texttt{lines}.
As a consequence, these programs are maximally verbose: they must compose many individual blocks to represent a \texttt{door}, let alone an entire \texttt{house}; and many individual lines to represent a polygon like a \texttt{hexagon}, let alone a complex \texttt{wheel}. 

To represent more complex shapes, we define higher-order graphics libraries that augment the initial set of base primitives with \textit{program subroutines} (Fig.~\ref{fig:library_gallery}) that encapsulate part structure (e.g., a subroutine for generating an entire \texttt{roof}).\footnote{Our approach to defining these higher-order libraries is analogous to the automated program library learning methods in \shortcite{ellis2020dreamcoder, tian2020learning, mccarthy2021learning, wang2021learning,wong2021leveraging}, which discover subroutines from a dataset containing programs that often correspond qualitatively to domain-relevant concepts.}
We constructed these libraries by abstracting out the nested, parametric functions used to generate each subdomain. 
In our experiments, we evaluate three libraries ($\mathcal{L}_1$, $\mathcal{L}_2$, and $\mathcal{L}_3$), each containing subroutines that build recursively on those at the previous level to yield increasingly complex visual parts.
For instance, $\mathcal{L}_1$ contains subroutines that abstract directly over the base library (e.g., from \texttt{lines} to \texttt{polygons}); and $\mathcal{L}_2$ contains subroutines that abstract additionally over those in $\mathcal{L}_1$ (e.g., \texttt{polygons} to \texttt{rings of polygons}). 
A given program $\pi_{\mathcal{L}_{base}}$ written in the base library can therefore be expressed equivalently---and more concisely---as $\pi_{\mathcal{L}_{i}}$ in one of the higher-order libraries. 
It is worth noting that higher-order libraries are thus defined \textit{cumulatively}:
$\mathcal{L}_1$ contains the new subroutines \textit{plus} the initial set of primitives in $\mathcal{L}_{base}$; and $\mathcal{L}_2$ contains even higher-order subroutines \textit{plus} all of the concepts in $\mathcal{L}_1$.

\paragraph{Modeling alignment between libraries and vocabularies}
For each subdomain, the set of libraries $\{\mathcal{L}_{base}, \mathcal{L}_1, \mathcal{L}_2, \mathcal{L}_3\}$ specifies a hypothesis space of alternative representations at differing levels of abstraction. We can now ask: which of these libraries best corresponds to the lexicon people use for each subdomain?
We formalize this notion of lexical correspondence with a \textit{library-to-vocabulary alignment metric} that reflects how closely the concepts in a given library co-occur with words across each subdomain. This metric is based a language-guided library learning model from the program synthesis literature \shortcite{wong2021leveraging}. In brief, we leverage a standard machine translation model, IBM Model 1 \shortcite{brown1993mathematics}, which can be fit to paired programs and instructions to estimate token-token translation probabilities $P(w|\rho)$ for each word $w \in W$ in the linguistic vocabulary and program component $\rho \in \mathcal{L}$ in the library. 
For each subdomain, we evaluate each library $\mathcal{L}_i$ using a cross-validation scheme (with batches of $n=5$ held out stimuli). 
We fit the model to all but the held-out stimuli and evaluate the \textit{mean per-word log-likelihood} for each held out instruction given its program in library $\mathcal{L}_i$. 
This metric varies monotonically as a function of negative {perplexity} \shortcite{wu2016google} and normalizes for instruction lengths. As in Part I, we consider only the \textit{what} phrases for each stimulus.


\begin{figure*}[ht!]
  \begin{center}
  \includegraphics[width=0.9\linewidth]{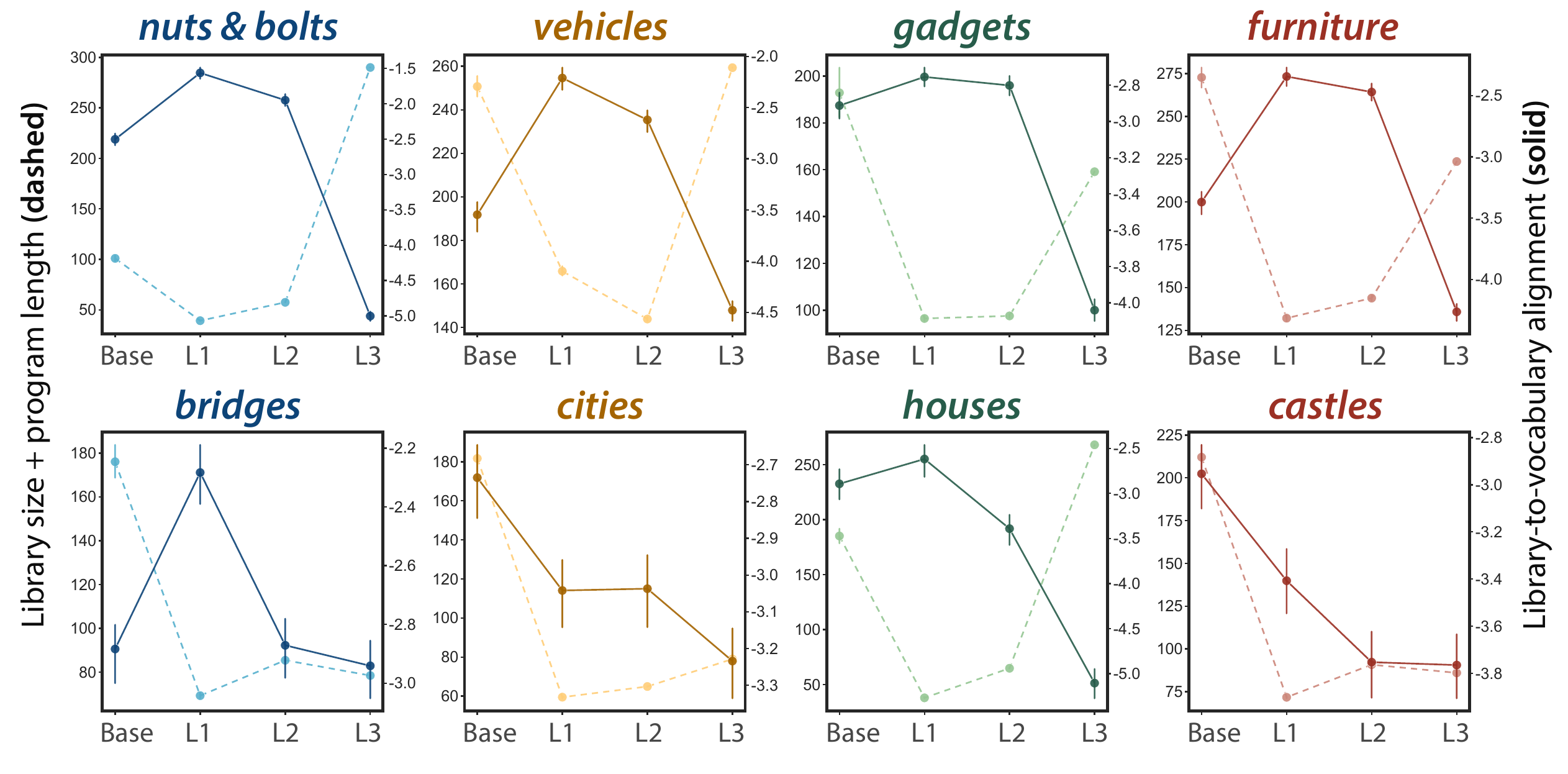}
  \caption{Relationship between concept libraries \{$L_{base}$, $\mathcal{L}_1$, $\mathcal{L}_2$, and $\mathcal{L}_3$\} (x-axis); combined library size and average program length in that library (dashed); and library-to-vocabulary alignment (solid).}\label{fig:perplexity-length}
  \end{center}
\end{figure*}

\subsection{Results}
\paragraph{Libraries produce different trade-off between concise object representation and overall library size} 
Supposing that any of these libraries captures the part concepts that people use when describing these objects, what would lead participants to favor one over another? Our hypothesis is that this choice reflects a trade-off between the value of compressing the length of programs $|\pi_{\mathcal{L}_i}|$ that represent individual objects and the value of reducing the total number of concepts $|\mathcal{L}_i|$ stored in the library (Fig. \ref{fig:library_gallery})\footnote{This trade-off between program description length $|\pi_{\mathcal{L}_i}|$ and library size $|\mathcal{L}_i|$ is described in greater detail in \shortcite{ellis2020dreamcoder} and analogous to the formulation in \shortcite{kirby2015compression}.}.
Higher-order libraries contain concepts that compress programs to a greater degree, as each program can be written by invoking a smaller number of more abstract subroutines.
However, each higher-order library is also larger than the last because it adds new concepts that must be represented along with all of the lower-level ones.



While library size increases monotonically with abstraction level, every subdomain has a non-monotonic \textit{combined representational cost} $C_{\mathcal{L}_i} = |\mathcal{L}_i| + \frac{1}{N} \sum_{\pi} |\pi_{\mathcal{L}_i}|$, where $N$ is the number of programs in the subdomain. 
A one-way ANOVA confirms that, in every subdomain, this combined cost measure systematically varies between libraries ($p$s $\ll$ 0.001), validating our assumption that these libraries capture different ways of negotiating the trade-off between object compression and library size. 
Further, as Fig. \ref{fig:perplexity-length} reveals, $C_{\mathcal{L}_i}$ (dashed line) typically follows a U-shaped curve. 
At the extremes, $C_{\mathcal{L}_{base}}$ is high because programs in $\mathcal{L}_{base}$ are verbose, whereas $C_{\mathcal{L}_3}$ is high due to the large size of ${\mathcal{L}_3}$. In all subdomains, $C_{\mathcal{L}_1}$ and $C_{\mathcal{L}_2}$ tend to be optimal because these intermediate libraries contain a set of useful part-based abstractions that capture recurring structure across many objects.

\paragraph{People favor vocabularies that jointly minimize object representation and library size}
We can now consider the results of our \textit{library-to-vocabulary alignment} model: which libraries best predict the words people use across each subdomain? 
To validate that this alignment metric is able to discriminate between libraries at all, we first conducted a one-way ANOVA on the alignment scores and found large and reliable differences between libraries in every subdomain ($p$s $<$ 0.001; Fig.~\ref{fig:perplexity-length}).

When we visualized these alignment scores (Fig. \ref{fig:perplexity-length}, solid lines), we observed that for the majority of the subdomains, the \textit{mean log-likelihoods} follows an inverted U-shaped curve.
Moreover, we generally find that the concept libraries that best predict language tend to be those containing parts of intermediate complexity --- for example, part concepts (e.g., individual \texttt{windows} or \texttt{wheels}) that lie between the lowest (e.g., \texttt{lines}) and highest (e.g., \texttt{hexagon with an inner ring of circular holes}) levels of abstraction in each domain.

Finally, we observed a striking correspondence between the libraries that optimize combined representational cost ($C_{\mathcal{L}_i}$) and those that score highly on their ability to predict language. 
This pattern, which held for most (though not all) subdomains, suggests that people generally prefer decomposing objects into nameable parts that can be reused for many objects across the full subdomain.

\subsection{Discussion}

The language we use to describe the world reveals the concepts with which we represent it.
In this paper, we look to natural language to investigate how people parse complex objects into meaningful parts --- for example, how people decompose a whole \texttt{train} into its \texttt{train cars} and \texttt{wheels}, or a \texttt{house} into its \texttt{windows}, \texttt{walls}, and \texttt{roof}.
We elicited descriptions for a large dataset of objects generated from \textit{graphics programs}, and present a computational approach for linking their generative and hierarchical structure with human descriptions. 
We find that the length of people's descriptions varies with the length of an object's generative program, establishing a basic correspondence between language and a program representations of object structure. 
By constructing higher-order \textit{concept libraries} which re-represent each object using more abstract program components, we find evidence that people's language reflects an underlying representational trade-off -- people prefer compact libraries of part concepts that efficiently capture structural motifs appearing in many objects. 
An intriguing implication of these findings is that there exists a ``basic level'' for part naming, by analogy to the well known basic level for object categories, and that can be explained by similar information-theoretic principles \shortcite{rosch1976basic}.

While these linguistic abstraction layers enable greater compression, they may also introduce downstream challenges for communication: terms with more abstract meanings may be less interpretable and/or too lossy in some cases (e.g., pedagogical contexts where learners may not be familiar with certain concepts).
To better understand how people communicate in these scenarios, it may be useful to conduct experiments manipulating what knowledge is shared between communicators to investigate the role of audience design and adaptation in interactive settings \shortcite{clark1982audience,krauss1991perspective,mccarthy2021learning}.

In other settings, the level of detail contained in the descriptions we collected may not be necessary to achieve certain communicative goals, such as object identification. 
A promising direction is to compare our descriptions to those produced in reference games where coarser distinctions between whole objects are sufficient, with the aim of understanding how task goals and context shape the \emph{relevance} of different levels of abstraction \shortcite{degen2020redundancy,bisk2020experience}. 

It is natural to expect substantial variation across descriptions in how well they support object understanding in others. To better understand why some descriptions are more informative than others, future work should also measure how well the descriptions we collected in the current study support the ability of other participants to accurately reconstruct the target objects.
Our approach and findings build on a recent and growing literature using programs \shortcite{lake2015human,goodman2014concepts} and libraries of functional components \shortcite{tian2020learning,mccarthy2021learning,wong2021leveraging} to model how people represent and communicate about the world. 
Our work generalizes previous insights into the statistical learning mechanisms that enable the rapid learning of visual regularities \shortcite{fiser2001unsupervised, orban2008bayesian, austerweil2013nonparametric} by proposing a more expressive program-like representation that can accommodate structure at multiple levels of abstraction.

More broadly, our work proposes and validates a general strategy for leveraging complex behavioral readouts (e.g., natural language descriptions) to draw rich and meaningful inferences about the content and structure of mental representations.
Such approaches have tremendous promise not only to advance cognitive theory, but may contribute to the design of artificial systems that learn more human-like abstractions.

\newpage

\bibliographystyle{apacite}

\setlength{\bibleftmargin}{.125in}
\setlength{\bibindent}{-\bibleftmargin}

\section{Acknowledgments}
RDH is supported by NSF grant
\#1911835 and a Princeton C.V. Starr Fellowship. JEF is supported by NSF CAREER \#2047191, an ONR Science of Autonomy award, and a Stanford Hoffman-Yee grant. GG is supported by an MIT Presidential Fellowship and an NSF Graduate Research Fellowship. CW, JBT, and JDA are supported by the MIT Quest for Intelligence, and CW and JBT have additional support from AFOSR Grant \#FA9550-19-1-0269,  the MIT-IBM Watson AI Lab, ONR Science of AI and DARPA Machine Common Sense.

\vspace{2em}
\fbox{\parbox[b][][c]{7.3cm}{\centering {All code and materials available at: \\
\href{https://github.com/cogtoolslab/lax-cogsci22}{\url{https://github.com/cogtoolslab/lax-cogsci22}}
}}}
\vspace{2em} \noindent

\bibliography{cogsci_2022}

\begin{thebibliography}{}

\bibitem [\protect \citeauthoryear {%
Austerweil%
\ \BBA {} Griffiths%
}{%
Austerweil%
\ \BBA {} Griffiths%
}{%
{\protect \APACyear {2013}}%
}]{%
austerweil2013nonparametric}
\APACinsertmetastar {%
austerweil2013nonparametric}%
\begin{APACrefauthors}%
Austerweil, J\BPBI L.%
\BCBT {}\ \BBA {} Griffiths, T\BPBI L.%
\end{APACrefauthors}%
\unskip\
\newblock
\APACrefYearMonthDay{2013}{}{}.
\newblock
{\BBOQ}\APACrefatitle {A nonparametric Bayesian framework for constructing
  flexible feature representations.} {A nonparametric bayesian framework for
  constructing flexible feature representations.}{\BBCQ}
\newblock
\APACjournalVolNumPages{Psychological Review}{120}{4}{817}.
\PrintBackRefs{\CurrentBib}

\bibitem [\protect \citeauthoryear {%
Bear%
\ \protect \BOthers {.}}{%
Bear%
\ \protect \BOthers {.}}{%
{\protect \APACyear {2020}}%
}]{%
bear2020learning}
\APACinsertmetastar {%
bear2020learning}%
\begin{APACrefauthors}%
Bear, D\BPBI M.%
, Fan, C.%
, Mrowca, D.%
, Li, Y.%
, Alter, S.%
, Nayebi, A.%
\BDBL {}others%
\end{APACrefauthors}%
\unskip\
\newblock
\APACrefYearMonthDay{2020}{}{}.
\newblock
{\BBOQ}\APACrefatitle {Learning physical graph representations from visual
  scenes} {Learning physical graph representations from visual scenes}.{\BBCQ}
\newblock
\APACjournalVolNumPages{Advances in Neural Information Processing
  Systems}{33}{}{}.
\PrintBackRefs{\CurrentBib}

\bibitem [\protect \citeauthoryear {%
Biederman%
}{%
Biederman%
}{%
{\protect \APACyear {1987}}%
}]{%
biederman1987recognition}
\APACinsertmetastar {%
biederman1987recognition}%
\begin{APACrefauthors}%
Biederman, I.%
\end{APACrefauthors}%
\unskip\
\newblock
\APACrefYearMonthDay{1987}{}{}.
\newblock
{\BBOQ}\APACrefatitle {Recognition-by-components: a theory of human image
  understanding.} {Recognition-by-components: a theory of human image
  understanding.}{\BBCQ}
\newblock
\APACjournalVolNumPages{Psychological Review}{94}{2}{115}.
\PrintBackRefs{\CurrentBib}

\bibitem [\protect \citeauthoryear {%
Bisk%
\ \protect \BOthers {.}}{%
Bisk%
\ \protect \BOthers {.}}{%
{\protect \APACyear {2020}}%
}]{%
bisk2020experience}
\APACinsertmetastar {%
bisk2020experience}%
\begin{APACrefauthors}%
Bisk, Y.%
, Holtzman, A.%
, Thomason, J.%
, Andreas, J.%
, Bengio, Y.%
, Chai, J.%
\BDBL {}others%
\end{APACrefauthors}%
\unskip\
\newblock
\APACrefYearMonthDay{2020}{}{}.
\newblock
{\BBOQ}\APACrefatitle {Experience grounds language} {Experience grounds
  language}.{\BBCQ}
\newblock
\APACjournalVolNumPages{arXiv preprint arXiv:2004.10151}{}{}{}.
\PrintBackRefs{\CurrentBib}

\bibitem [\protect \citeauthoryear {%
Brown%
, Della~Pietra%
, Della~Pietra%
\BCBL {}\ \BBA {} Mercer%
}{%
Brown%
\ \protect \BOthers {.}}{%
{\protect \APACyear {1993}}%
}]{%
brown1993mathematics}
\APACinsertmetastar {%
brown1993mathematics}%
\begin{APACrefauthors}%
Brown, P\BPBI F.%
, Della~Pietra, S\BPBI A.%
, Della~Pietra, V\BPBI J.%
\BCBL {}\ \BBA {} Mercer, R\BPBI L.%
\end{APACrefauthors}%
\unskip\
\newblock
\APACrefYearMonthDay{1993}{}{}.
\newblock
{\BBOQ}\APACrefatitle {The Mathematics of Statistical Machine Translation:
  Parameter Estimation} {The mathematics of statistical machine translation:
  Parameter estimation}.{\BBCQ}
\newblock
\APACjournalVolNumPages{Computational Linguistics}{19}{2}{263--311}.
\PrintBackRefs{\CurrentBib}

\bibitem [\protect \citeauthoryear {%
Clark%
\ \BBA {} Murphy%
}{%
Clark%
\ \BBA {} Murphy%
}{%
{\protect \APACyear {1982}}%
}]{%
clark1982audience}
\APACinsertmetastar {%
clark1982audience}%
\begin{APACrefauthors}%
Clark, H\BPBI H.%
\BCBT {}\ \BBA {} Murphy, G\BPBI L.%
\end{APACrefauthors}%
\unskip\
\newblock
\APACrefYearMonthDay{1982}{}{}.
\newblock
{\BBOQ}\APACrefatitle {Audience design in meaning and reference} {Audience
  design in meaning and reference}.{\BBCQ}
\newblock
\BIn{} \APACrefbtitle {Advances in psychology} {Advances in psychology}\
  (\BVOL~9, \BPGS\ 287--299).
\newblock
\APACaddressPublisher{}{Elsevier}.
\PrintBackRefs{\CurrentBib}

\bibitem [\protect \citeauthoryear {%
Degen%
, Hawkins%
, Graf%
, Kreiss%
\BCBL {}\ \BBA {} Goodman%
}{%
Degen%
\ \protect \BOthers {.}}{%
{\protect \APACyear {2020}}%
}]{%
degen2020redundancy}
\APACinsertmetastar {%
degen2020redundancy}%
\begin{APACrefauthors}%
Degen, J.%
, Hawkins, R\BPBI D.%
, Graf, C.%
, Kreiss, E.%
\BCBL {}\ \BBA {} Goodman, N\BPBI D.%
\end{APACrefauthors}%
\unskip\
\newblock
\APACrefYearMonthDay{2020}{}{}.
\newblock
{\BBOQ}\APACrefatitle {When redundancy is useful: A Bayesian approach to
  ``overinformative'' referring expressions.} {When redundancy is useful: A
  bayesian approach to ``overinformative'' referring expressions.}{\BBCQ}
\newblock
\APACjournalVolNumPages{Psychological Review}{127}{4}{591}.
\PrintBackRefs{\CurrentBib}

\bibitem [\protect \citeauthoryear {%
Ellis%
\ \protect \BOthers {.}}{%
Ellis%
\ \protect \BOthers {.}}{%
{\protect \APACyear {2020}}%
}]{%
ellis2020dreamcoder}
\APACinsertmetastar {%
ellis2020dreamcoder}%
\begin{APACrefauthors}%
Ellis, K.%
, Wong, C.%
, Nye, M.%
, Sable-Meyer, M.%
, Cary, L.%
, Morales, L.%
\BDBL {}Tenenbaum, J\BPBI B.%
\end{APACrefauthors}%
\unskip\
\newblock
\APACrefYearMonthDay{2020}{}{}.
\newblock
{\BBOQ}\APACrefatitle {Dreamcoder: Growing generalizable, interpretable
  knowledge with wake-sleep bayesian program learning} {Dreamcoder: Growing
  generalizable, interpretable knowledge with wake-sleep bayesian program
  learning}.{\BBCQ}
\newblock
\APACjournalVolNumPages{arXiv preprint arXiv:2006.08381}{}{}{}.
\PrintBackRefs{\CurrentBib}

\bibitem [\protect \citeauthoryear {%
Fiser%
\ \BBA {} Aslin%
}{%
Fiser%
\ \BBA {} Aslin%
}{%
{\protect \APACyear {2001}}%
}]{%
fiser2001unsupervised}
\APACinsertmetastar {%
fiser2001unsupervised}%
\begin{APACrefauthors}%
Fiser, J.%
\BCBT {}\ \BBA {} Aslin, R\BPBI N.%
\end{APACrefauthors}%
\unskip\
\newblock
\APACrefYearMonthDay{2001}{}{}.
\newblock
{\BBOQ}\APACrefatitle {Unsupervised statistical learning of higher-order
  spatial structures from visual scenes} {Unsupervised statistical learning of
  higher-order spatial structures from visual scenes}.{\BBCQ}
\newblock
\APACjournalVolNumPages{Psychological Science}{12}{6}{499--504}.
\PrintBackRefs{\CurrentBib}

\bibitem [\protect \citeauthoryear {%
Goodman%
, Tenenbaum%
\BCBL {}\ \BBA {} Gerstenberg%
}{%
Goodman%
\ \protect \BOthers {.}}{%
{\protect \APACyear {2014}}%
}]{%
goodman2014concepts}
\APACinsertmetastar {%
goodman2014concepts}%
\begin{APACrefauthors}%
Goodman, N\BPBI D.%
, Tenenbaum, J\BPBI B.%
\BCBL {}\ \BBA {} Gerstenberg, T.%
\end{APACrefauthors}%
\unskip\
\newblock
\APACrefYearMonthDay{2014}{}{}.
\newblock
{\BBOQ}\APACrefatitle {Concepts in a probabilistic language of thought}
  {Concepts in a probabilistic language of thought}.{\BBCQ}
\newblock
\BIn{} \APACrefbtitle {The Conceptual Mind: New Directions in the Study of
  Concepts.} {The conceptual mind: New directions in the study of concepts.}
\PrintBackRefs{\CurrentBib}

\bibitem [\protect \citeauthoryear {%
Hawkins%
, Frank%
\BCBL {}\ \BBA {} Goodman%
}{%
Hawkins%
\ \protect \BOthers {.}}{%
{\protect \APACyear {2020}}%
}]{%
hawkins2020characterizing}
\APACinsertmetastar {%
hawkins2020characterizing}%
\begin{APACrefauthors}%
Hawkins, R\BPBI D.%
, Frank, M\BPBI C.%
\BCBL {}\ \BBA {} Goodman, N\BPBI D.%
\end{APACrefauthors}%
\unskip\
\newblock
\APACrefYearMonthDay{2020}{}{}.
\newblock
{\BBOQ}\APACrefatitle {Characterizing the dynamics of learning in repeated
  reference games} {Characterizing the dynamics of learning in repeated
  reference games}.{\BBCQ}
\newblock
\APACjournalVolNumPages{Cognitive Science}{44}{6}{e12845}.
\PrintBackRefs{\CurrentBib}

\bibitem [\protect \citeauthoryear {%
Hoffman%
\ \BBA {} Richards%
}{%
Hoffman%
\ \BBA {} Richards%
}{%
{\protect \APACyear {1984}}%
}]{%
hoffman1984parts}
\APACinsertmetastar {%
hoffman1984parts}%
\begin{APACrefauthors}%
Hoffman, D\BPBI D.%
\BCBT {}\ \BBA {} Richards, W\BPBI A.%
\end{APACrefauthors}%
\unskip\
\newblock
\APACrefYearMonthDay{1984}{}{}.
\newblock
{\BBOQ}\APACrefatitle {Parts of recognition} {Parts of recognition}.{\BBCQ}
\newblock
\APACjournalVolNumPages{Cognition}{18}{1-3}{65--96}.
\PrintBackRefs{\CurrentBib}

\bibitem [\protect \citeauthoryear {%
Kirby%
, Tamariz%
, Cornish%
\BCBL {}\ \BBA {} Smith%
}{%
Kirby%
\ \protect \BOthers {.}}{%
{\protect \APACyear {2015}}%
}]{%
kirby2015compression}
\APACinsertmetastar {%
kirby2015compression}%
\begin{APACrefauthors}%
Kirby, S.%
, Tamariz, M.%
, Cornish, H.%
\BCBL {}\ \BBA {} Smith, K.%
\end{APACrefauthors}%
\unskip\
\newblock
\APACrefYearMonthDay{2015}{}{}.
\newblock
{\BBOQ}\APACrefatitle {Compression and communication in the cultural evolution
  of linguistic structure} {Compression and communication in the cultural
  evolution of linguistic structure}.{\BBCQ}
\newblock
\APACjournalVolNumPages{Cognition}{141}{}{87--102}.
\PrintBackRefs{\CurrentBib}

\bibitem [\protect \citeauthoryear {%
Krauss%
\ \BBA {} Fussell%
}{%
Krauss%
\ \BBA {} Fussell%
}{%
{\protect \APACyear {1991}}%
}]{%
krauss1991perspective}
\APACinsertmetastar {%
krauss1991perspective}%
\begin{APACrefauthors}%
Krauss, R\BPBI M.%
\BCBT {}\ \BBA {} Fussell, S\BPBI R.%
\end{APACrefauthors}%
\unskip\
\newblock
\APACrefYearMonthDay{1991}{}{}.
\newblock
{\BBOQ}\APACrefatitle {Perspective-taking in communication: Representations of
  others' knowledge in reference} {Perspective-taking in communication:
  Representations of others' knowledge in reference}.{\BBCQ}
\newblock
\APACjournalVolNumPages{Social Cognition}{9}{1}{2--24}.
\PrintBackRefs{\CurrentBib}

\bibitem [\protect \citeauthoryear {%
Lake%
, Salakhutdinov%
\BCBL {}\ \BBA {} Tenenbaum%
}{%
Lake%
\ \protect \BOthers {.}}{%
{\protect \APACyear {2015}}%
}]{%
lake2015human}
\APACinsertmetastar {%
lake2015human}%
\begin{APACrefauthors}%
Lake, B\BPBI M.%
, Salakhutdinov, R.%
\BCBL {}\ \BBA {} Tenenbaum, J\BPBI B.%
\end{APACrefauthors}%
\unskip\
\newblock
\APACrefYearMonthDay{2015}{}{}.
\newblock
{\BBOQ}\APACrefatitle {Human-level concept learning through probabilistic
  program induction} {Human-level concept learning through probabilistic
  program induction}.{\BBCQ}
\newblock
\APACjournalVolNumPages{Science}{350}{6266}{1332--1338}.
\PrintBackRefs{\CurrentBib}

\bibitem [\protect \citeauthoryear {%
Landau%
\ \BBA {} Jackendoff%
}{%
Landau%
\ \BBA {} Jackendoff%
}{%
{\protect \APACyear {1993}}%
}]{%
landau1993and}
\APACinsertmetastar {%
landau1993and}%
\begin{APACrefauthors}%
Landau, B.%
\BCBT {}\ \BBA {} Jackendoff, R.%
\end{APACrefauthors}%
\unskip\
\newblock
\APACrefYearMonthDay{1993}{}{}.
\newblock
{\BBOQ}\APACrefatitle {``{W}hat'' and ``where'' in spatial language and spatial
  cognition} {``{W}hat'' and ``where'' in spatial language and spatial
  cognition}.{\BBCQ}
\newblock
\APACjournalVolNumPages{Behavioral and Brain Sciences}{16}{2}{}.
\PrintBackRefs{\CurrentBib}

\bibitem [\protect \citeauthoryear {%
Markman%
\ \BBA {} Wachtel%
}{%
Markman%
\ \BBA {} Wachtel%
}{%
{\protect \APACyear {1988}}%
}]{%
markman1988children}
\APACinsertmetastar {%
markman1988children}%
\begin{APACrefauthors}%
Markman, E\BPBI M.%
\BCBT {}\ \BBA {} Wachtel, G\BPBI F.%
\end{APACrefauthors}%
\unskip\
\newblock
\APACrefYearMonthDay{1988}{}{}.
\newblock
{\BBOQ}\APACrefatitle {Children's use of mutual exclusivity to constrain the
  meanings of words} {Children's use of mutual exclusivity to constrain the
  meanings of words}.{\BBCQ}
\newblock
\APACjournalVolNumPages{Cognitive Psychology}{20}{2}{121--157}.
\PrintBackRefs{\CurrentBib}

\bibitem [\protect \citeauthoryear {%
Marr%
\ \BBA {} Nishihara%
}{%
Marr%
\ \BBA {} Nishihara%
}{%
{\protect \APACyear {1978}}%
}]{%
marr1978representation}
\APACinsertmetastar {%
marr1978representation}%
\begin{APACrefauthors}%
Marr, D.%
\BCBT {}\ \BBA {} Nishihara, H\BPBI K.%
\end{APACrefauthors}%
\unskip\
\newblock
\APACrefYearMonthDay{1978}{}{}.
\newblock
{\BBOQ}\APACrefatitle {Representation and recognition of the spatial
  organization of three-dimensional shapes} {Representation and recognition of
  the spatial organization of three-dimensional shapes}.{\BBCQ}
\newblock
\APACjournalVolNumPages{Proceedings of the Royal Society of London. Series B.
  Biological Sciences}{200}{1140}{269--294}.
\PrintBackRefs{\CurrentBib}

\bibitem [\protect \citeauthoryear {%
McCarthy%
, Hawkins%
, Wang%
, Holdaway%
\BCBL {}\ \BBA {} Fan%
}{%
McCarthy%
\ \protect \BOthers {.}}{%
{\protect \APACyear {2021}}%
}]{%
mccarthy2021learning}
\APACinsertmetastar {%
mccarthy2021learning}%
\begin{APACrefauthors}%
McCarthy, W\BPBI P.%
, Hawkins, R\BPBI D.%
, Wang, H.%
, Holdaway, C.%
\BCBL {}\ \BBA {} Fan, J\BPBI E.%
\end{APACrefauthors}%
\unskip\
\newblock
\APACrefYearMonthDay{2021}{}{}.
\newblock
{\BBOQ}\APACrefatitle {Learning to communicate about shared procedural
  abstractions} {Learning to communicate about shared procedural
  abstractions}.{\BBCQ}
\newblock
\BIn{} \APACrefbtitle {Proceedings of the Annual Meeting of the Cognitive
  Science Society} {Proceedings of the annual meeting of the cognitive science
  society}\ (\BPGS\ 77--83).
\PrintBackRefs{\CurrentBib}

\bibitem [\protect \citeauthoryear {%
Miller%
\ \BBA {} Johnson-Laird%
}{%
Miller%
\ \BBA {} Johnson-Laird%
}{%
{\protect \APACyear {1976}}%
}]{%
miller2013language}
\APACinsertmetastar {%
miller2013language}%
\begin{APACrefauthors}%
Miller, G\BPBI A.%
\BCBT {}\ \BBA {} Johnson-Laird, P\BPBI N.%
\end{APACrefauthors}%
\unskip\
\newblock
\APACrefYear{1976}.
\newblock
\APACrefbtitle {Language and perception} {Language and perception}.
\newblock
\APACaddressPublisher{}{Harvard University Press}.
\PrintBackRefs{\CurrentBib}

\bibitem [\protect \citeauthoryear {%
Mo%
\ \protect \BOthers {.}}{%
Mo%
\ \protect \BOthers {.}}{%
{\protect \APACyear {2019}}%
}]{%
mo2019partnet}
\APACinsertmetastar {%
mo2019partnet}%
\begin{APACrefauthors}%
Mo, K.%
, Zhu, S.%
, Chang, A\BPBI X.%
, Yi, L.%
, Tripathi, S.%
, Guibas, L\BPBI J.%
\BCBL {}\ \BBA {} Su, H.%
\end{APACrefauthors}%
\unskip\
\newblock
\APACrefYearMonthDay{2019}{}{}.
\newblock
{\BBOQ}\APACrefatitle {Partnet: A large-scale benchmark for fine-grained and
  hierarchical part-level 3d object understanding} {Partnet: A large-scale
  benchmark for fine-grained and hierarchical part-level 3d object
  understanding}.{\BBCQ}
\newblock
\BIn{} \APACrefbtitle {Proceedings of the {IEEE/CVF} {C}onference on {C}omputer
  {V}ision and {P}attern {R}ecognition} {Proceedings of the {IEEE/CVF}
  {C}onference on {C}omputer {V}ision and {P}attern {R}ecognition}\ (\BPGS\
  909--918).
\PrintBackRefs{\CurrentBib}

\bibitem [\protect \citeauthoryear {%
Mukherjee%
, Hawkins%
\BCBL {}\ \BBA {} Fan%
}{%
Mukherjee%
\ \protect \BOthers {.}}{%
{\protect \APACyear {2019}}%
}]{%
mukherjee2019communicating}
\APACinsertmetastar {%
mukherjee2019communicating}%
\begin{APACrefauthors}%
Mukherjee, K.%
, Hawkins, R\BPBI D.%
\BCBL {}\ \BBA {} Fan, J\BPBI W.%
\end{APACrefauthors}%
\unskip\
\newblock
\APACrefYearMonthDay{2019}{}{}.
\newblock
{\BBOQ}\APACrefatitle {Communicating semantic part information in drawings.}
  {Communicating semantic part information in drawings.}{\BBCQ}
\newblock
\BIn{} \APACrefbtitle {Proceedings of the {A}nnual {M}eeting of the {C}ognitive
  {S}cience {S}ociety} {Proceedings of the {A}nnual {M}eeting of the
  {C}ognitive {S}cience {S}ociety}\ (\BPGS\ 2413--2419).
\PrintBackRefs{\CurrentBib}

\bibitem [\protect \citeauthoryear {%
Orb{\'a}n%
, Fiser%
, Aslin%
\BCBL {}\ \BBA {} Lengyel%
}{%
Orb{\'a}n%
\ \protect \BOthers {.}}{%
{\protect \APACyear {2008}}%
}]{%
orban2008bayesian}
\APACinsertmetastar {%
orban2008bayesian}%
\begin{APACrefauthors}%
Orb{\'a}n, G.%
, Fiser, J.%
, Aslin, R\BPBI N.%
\BCBL {}\ \BBA {} Lengyel, M.%
\end{APACrefauthors}%
\unskip\
\newblock
\APACrefYearMonthDay{2008}{}{}.
\newblock
{\BBOQ}\APACrefatitle {Bayesian learning of visual chunks by human observers}
  {Bayesian learning of visual chunks by human observers}.{\BBCQ}
\newblock
\APACjournalVolNumPages{Proceedings of the National Academy of
  Sciences}{105}{7}{2745--2750}.
\PrintBackRefs{\CurrentBib}

\bibitem [\protect \citeauthoryear {%
Palmer%
}{%
Palmer%
}{%
{\protect \APACyear {1977}}%
}]{%
palmer1977hierarchical}
\APACinsertmetastar {%
palmer1977hierarchical}%
\begin{APACrefauthors}%
Palmer, S\BPBI E.%
\end{APACrefauthors}%
\unskip\
\newblock
\APACrefYearMonthDay{1977}{}{}.
\newblock
{\BBOQ}\APACrefatitle {Hierarchical structure in perceptual representation}
  {Hierarchical structure in perceptual representation}.{\BBCQ}
\newblock
\APACjournalVolNumPages{Cognitive Psychology}{9}{4}{441--474}.
\PrintBackRefs{\CurrentBib}

\bibitem [\protect \citeauthoryear {%
Regier%
, Kemp%
\BCBL {}\ \BBA {} Kay%
}{%
Regier%
\ \protect \BOthers {.}}{%
{\protect \APACyear {2015}}%
}]{%
regier201511}
\APACinsertmetastar {%
regier201511}%
\begin{APACrefauthors}%
Regier, T.%
, Kemp, C.%
\BCBL {}\ \BBA {} Kay, P.%
\end{APACrefauthors}%
\unskip\
\newblock
\APACrefYearMonthDay{2015}{}{}.
\newblock
{\BBOQ}\APACrefatitle {Word Meanings across Languages Support Efficient
  Communication} {Word meanings across languages support efficient
  communication}.{\BBCQ}
\newblock
\APACjournalVolNumPages{The handbook of language emergence}{87}{}{237}.
\PrintBackRefs{\CurrentBib}

\bibitem [\protect \citeauthoryear {%
Rosch%
\ \BBA {} Mervis%
}{%
Rosch%
\ \BBA {} Mervis%
}{%
{\protect \APACyear {1975}}%
}]{%
rosch1975family}
\APACinsertmetastar {%
rosch1975family}%
\begin{APACrefauthors}%
Rosch, E.%
\BCBT {}\ \BBA {} Mervis, C\BPBI B.%
\end{APACrefauthors}%
\unskip\
\newblock
\APACrefYearMonthDay{1975}{}{}.
\newblock
{\BBOQ}\APACrefatitle {Family resemblances: Studies in the internal structure
  of categories} {Family resemblances: Studies in the internal structure of
  categories}.{\BBCQ}
\newblock
\APACjournalVolNumPages{Cognitive Psychology}{7}{4}{573--605}.
\PrintBackRefs{\CurrentBib}

\bibitem [\protect \citeauthoryear {%
Rosch%
, Mervis%
, Gray%
, Johnson%
\BCBL {}\ \BBA {} Boyes-Braem%
}{%
Rosch%
\ \protect \BOthers {.}}{%
{\protect \APACyear {1976}}%
}]{%
rosch1976basic}
\APACinsertmetastar {%
rosch1976basic}%
\begin{APACrefauthors}%
Rosch, E.%
, Mervis, C\BPBI B.%
, Gray, W\BPBI D.%
, Johnson, D\BPBI M.%
\BCBL {}\ \BBA {} Boyes-Braem, P.%
\end{APACrefauthors}%
\unskip\
\newblock
\APACrefYearMonthDay{1976}{}{}.
\newblock
{\BBOQ}\APACrefatitle {Basic objects in natural categories} {Basic objects in
  natural categories}.{\BBCQ}
\newblock
\APACjournalVolNumPages{Cognitive psychology}{8}{3}{382--439}.
\PrintBackRefs{\CurrentBib}

\bibitem [\protect \citeauthoryear {%
Schyns%
\ \BBA {} Murphy%
}{%
Schyns%
\ \BBA {} Murphy%
}{%
{\protect \APACyear {1994}}%
}]{%
schyns1994ontogeny}
\APACinsertmetastar {%
schyns1994ontogeny}%
\begin{APACrefauthors}%
Schyns, P\BPBI G.%
\BCBT {}\ \BBA {} Murphy, G\BPBI L.%
\end{APACrefauthors}%
\unskip\
\newblock
\APACrefYearMonthDay{1994}{}{}.
\newblock
{\BBOQ}\APACrefatitle {The ontogeny of part representation in object concepts}
  {The ontogeny of part representation in object concepts}.{\BBCQ}
\newblock
\APACjournalVolNumPages{The Psychology of Learning and
  Motivation}{31}{}{305--349}.
\PrintBackRefs{\CurrentBib}

\bibitem [\protect \citeauthoryear {%
Sun%
\ \BBA {} Firestone%
}{%
Sun%
\ \BBA {} Firestone%
}{%
{\protect \APACyear {2021}}%
}]{%
sun2021seeing}
\APACinsertmetastar {%
sun2021seeing}%
\begin{APACrefauthors}%
Sun, Z.%
\BCBT {}\ \BBA {} Firestone, C.%
\end{APACrefauthors}%
\unskip\
\newblock
\APACrefYearMonthDay{2021}{}{}.
\newblock
{\BBOQ}\APACrefatitle {Seeing and speaking: How verbal “description length”
  encodes visual complexity.} {Seeing and speaking: How verbal “description
  length” encodes visual complexity.}{\BBCQ}
\newblock
\APACjournalVolNumPages{Journal of Experimental Psychology:
  General}{151}{1}{82–96}.
\PrintBackRefs{\CurrentBib}

\bibitem [\protect \citeauthoryear {%
Tian%
, Ellis%
, Kryven%
\BCBL {}\ \BBA {} Tenenbaum%
}{%
Tian%
\ \protect \BOthers {.}}{%
{\protect \APACyear {2020}}%
}]{%
tian2020learning}
\APACinsertmetastar {%
tian2020learning}%
\begin{APACrefauthors}%
Tian, L.%
, Ellis, K.%
, Kryven, M.%
\BCBL {}\ \BBA {} Tenenbaum, J.%
\end{APACrefauthors}%
\unskip\
\newblock
\APACrefYearMonthDay{2020}{}{}.
\newblock
{\BBOQ}\APACrefatitle {Learning abstract structure for drawing by efficient
  motor program induction} {Learning abstract structure for drawing by
  efficient motor program induction}.{\BBCQ}
\newblock
\APACjournalVolNumPages{Advances in Neural Information Processing
  Systems}{33}{}{}.
\PrintBackRefs{\CurrentBib}

\bibitem [\protect \citeauthoryear {%
Tversky%
}{%
Tversky%
}{%
{\protect \APACyear {1989}}%
}]{%
tversky1989parts}
\APACinsertmetastar {%
tversky1989parts}%
\begin{APACrefauthors}%
Tversky, B.%
\end{APACrefauthors}%
\unskip\
\newblock
\APACrefYearMonthDay{1989}{}{}.
\newblock
{\BBOQ}\APACrefatitle {Parts, partonomies, and taxonomies.} {Parts,
  partonomies, and taxonomies.}{\BBCQ}
\newblock
\APACjournalVolNumPages{Developmental Psychology}{25}{6}{983}.
\PrintBackRefs{\CurrentBib}

\bibitem [\protect \citeauthoryear {%
Tversky%
\ \BBA {} Hemenway%
}{%
Tversky%
\ \BBA {} Hemenway%
}{%
{\protect \APACyear {1984}}%
}]{%
tversky1984objects}
\APACinsertmetastar {%
tversky1984objects}%
\begin{APACrefauthors}%
Tversky, B.%
\BCBT {}\ \BBA {} Hemenway, K.%
\end{APACrefauthors}%
\unskip\
\newblock
\APACrefYearMonthDay{1984}{}{}.
\newblock
{\BBOQ}\APACrefatitle {Objects, parts, and categories.} {Objects, parts, and
  categories.}{\BBCQ}
\newblock
\APACjournalVolNumPages{Journal of Experimental Psychology:
  General}{113}{2}{169}.
\PrintBackRefs{\CurrentBib}

\bibitem [\protect \citeauthoryear {%
Wang%
, Polikarpova%
\BCBL {}\ \BBA {} Fan%
}{%
Wang%
\ \protect \BOthers {.}}{%
{\protect \APACyear {2021}}%
}]{%
wang2021learning}
\APACinsertmetastar {%
wang2021learning}%
\begin{APACrefauthors}%
Wang, H.%
, Polikarpova, N.%
\BCBL {}\ \BBA {} Fan, J\BPBI E.%
\end{APACrefauthors}%
\unskip\
\newblock
\APACrefYearMonthDay{2021}{}{}.
\newblock
{\BBOQ}\APACrefatitle {Learning part-based abstractions for visual object
  concepts} {Learning part-based abstractions for visual object
  concepts}.{\BBCQ}
\newblock
\BIn{} \APACrefbtitle {Proceedings of the {A}nnual {M}eeting of the {C}ognitive
  {S}cience {S}ociety.} {Proceedings of the {A}nnual {M}eeting of the
  {C}ognitive {S}cience {S}ociety.}
\PrintBackRefs{\CurrentBib}

\bibitem [\protect \citeauthoryear {%
Wong%
, Ellis%
, Tenenbaum%
\BCBL {}\ \BBA {} Andreas%
}{%
Wong%
\ \protect \BOthers {.}}{%
{\protect \APACyear {2021}}%
}]{%
wong2021leveraging}
\APACinsertmetastar {%
wong2021leveraging}%
\begin{APACrefauthors}%
Wong, C.%
, Ellis, K\BPBI M.%
, Tenenbaum, J.%
\BCBL {}\ \BBA {} Andreas, J.%
\end{APACrefauthors}%
\unskip\
\newblock
\APACrefYearMonthDay{2021}{}{}.
\newblock
{\BBOQ}\APACrefatitle {Leveraging language to learn program abstractions and
  search heuristics} {Leveraging language to learn program abstractions and
  search heuristics}.{\BBCQ}
\newblock
\BIn{} \APACrefbtitle {International {C}onference on {M}achine {L}earning}
  {International {C}onference on {M}achine {L}earning}\ (\BPGS\ 11193--11204).
\PrintBackRefs{\CurrentBib}

\bibitem [\protect \citeauthoryear {%
Wu%
\ \protect \BOthers {.}}{%
Wu%
\ \protect \BOthers {.}}{%
{\protect \APACyear {2016}}%
}]{%
wu2016google}
\APACinsertmetastar {%
wu2016google}%
\begin{APACrefauthors}%
Wu, Y.%
, Schuster, M.%
, Chen, Z.%
, Le, Q\BPBI V.%
, Norouzi, M.%
, Macherey, W.%
\BDBL {}others%
\end{APACrefauthors}%
\unskip\
\newblock
\APACrefYearMonthDay{2016}{}{}.
\newblock
{\BBOQ}\APACrefatitle {Google's neural machine translation system: Bridging the
  gap between human and machine translation} {Google's neural machine
  translation system: Bridging the gap between human and machine
  translation}.{\BBCQ}
\newblock
\APACjournalVolNumPages{arXiv preprint arXiv:1609.08144}{}{}{}.
\PrintBackRefs{\CurrentBib}

\bibitem [\protect \citeauthoryear {%
Zaslavsky%
, Kemp%
, Regier%
\BCBL {}\ \BBA {} Tishby%
}{%
Zaslavsky%
\ \protect \BOthers {.}}{%
{\protect \APACyear {2018}}%
}]{%
zaslavsky2018efficient}
\APACinsertmetastar {%
zaslavsky2018efficient}%
\begin{APACrefauthors}%
Zaslavsky, N.%
, Kemp, C.%
, Regier, T.%
\BCBL {}\ \BBA {} Tishby, N.%
\end{APACrefauthors}%
\unskip\
\newblock
\APACrefYearMonthDay{2018}{}{}.
\newblock
{\BBOQ}\APACrefatitle {Efficient compression in color naming and its evolution}
  {Efficient compression in color naming and its evolution}.{\BBCQ}
\newblock
\APACjournalVolNumPages{Proceedings of the National Academy of
  Sciences}{115}{31}{7937--7942}.
\PrintBackRefs{\CurrentBib}

\end{thebibliography}


\begin{thebibliography}{}

\bibitem [\protect \citeauthoryear {%
Brown%
, Della~Pietra%
, Della~Pietra%
\BCBL {}\ \BBA {} Mercer%
}{%
Brown%
\ \protect \BOthers {.}}{%
{\protect \APACyear {1993}}%
}]{%
brown1993mathematics}
\APACinsertmetastar {%
brown1993mathematics}%
\begin{APACrefauthors}%
Brown, P\BPBI F.%
, Della~Pietra, S\BPBI A.%
, Della~Pietra, V\BPBI J.%
\BCBL {}\ \BBA {} Mercer, R\BPBI L.%
\end{APACrefauthors}%
\unskip\
\newblock
\APACrefYearMonthDay{1993}{}{}.
\newblock
{\BBOQ}\APACrefatitle {The Mathematics of Statistical Machine Translation:
  Parameter Estimation} {The mathematics of statistical machine translation:
  Parameter estimation}.{\BBCQ}
\newblock
\APACjournalVolNumPages{Computational Linguistics}{19}{2}{263--311}.
\PrintBackRefs{\CurrentBib}

\bibitem [\protect \citeauthoryear {%
Ellis%
\ \protect \BOthers {.}}{%
Ellis%
\ \protect \BOthers {.}}{%
{\protect \APACyear {2020}}%
}]{%
ellis2020dreamcoder}
\APACinsertmetastar {%
ellis2020dreamcoder}%
\begin{APACrefauthors}%
Ellis, K.%
, Wong, C.%
, Nye, M.%
, Sable-Meyer, M.%
, Cary, L.%
, Morales, L.%
\BDBL {}Tenenbaum, J\BPBI B.%
\end{APACrefauthors}%
\unskip\
\newblock
\APACrefYearMonthDay{2020}{}{}.
\newblock
{\BBOQ}\APACrefatitle {Dreamcoder: Growing generalizable, interpretable
  knowledge with wake-sleep bayesian program learning} {Dreamcoder: Growing
  generalizable, interpretable knowledge with wake-sleep bayesian program
  learning}.{\BBCQ}
\newblock
\APACjournalVolNumPages{arXiv preprint arXiv:2006.08381}{}{}{}.
\PrintBackRefs{\CurrentBib}

\bibitem [\protect \citeauthoryear {%
Wang%
, Polikarpova%
\BCBL {}\ \BBA {} Fan%
}{%
Wang%
\ \protect \BOthers {.}}{%
{\protect \APACyear {2021}}%
}]{%
wang2021learning}
\APACinsertmetastar {%
wang2021learning}%
\begin{APACrefauthors}%
Wang, H.%
, Polikarpova, N.%
\BCBL {}\ \BBA {} Fan, J\BPBI E.%
\end{APACrefauthors}%
\unskip\
\newblock
\APACrefYearMonthDay{2021}{}{}.
\newblock
{\BBOQ}\APACrefatitle {Learning part-based abstractions for visual object
  concepts} {Learning part-based abstractions for visual object
  concepts}.{\BBCQ}
\newblock
\BIn{} \APACrefbtitle {Proceedings of the {A}nnual {M}eeting of the {C}ognitive
  {S}cience {S}ociety.} {Proceedings of the {A}nnual {M}eeting of the
  {C}ognitive {S}cience {S}ociety.}
\PrintBackRefs{\CurrentBib}

\bibitem [\protect \citeauthoryear {%
Wong%
, Ellis%
, Tenenbaum%
\BCBL {}\ \BBA {} Andreas%
}{%
Wong%
\ \protect \BOthers {.}}{%
{\protect \APACyear {2021}}%
}]{%
wong2021leveraging}
\APACinsertmetastar {%
wong2021leveraging}%
\begin{APACrefauthors}%
Wong, C.%
, Ellis, K\BPBI M.%
, Tenenbaum, J.%
\BCBL {}\ \BBA {} Andreas, J.%
\end{APACrefauthors}%
\unskip\
\newblock
\APACrefYearMonthDay{2021}{}{}.
\newblock
{\BBOQ}\APACrefatitle {Leveraging language to learn program abstractions and
  search heuristics} {Leveraging language to learn program abstractions and
  search heuristics}.{\BBCQ}
\newblock
\BIn{} \APACrefbtitle {International {C}onference on {M}achine {L}earning}
  {International {C}onference on {M}achine {L}earning}\ (\BPGS\ 11193--11204).
\PrintBackRefs{\CurrentBib}

\end{thebibliography}

\end{document}


\maketitle

\section{S1. Part I Supplemental Details} \label{sec:s1_part_i}
This section contains additional details on the stimulus generation procedure in \textbf{Part I} of the main paper.

\subsection{Stimulus generation}
As discussed in the main paper, the dataset comprises two distinct stimulus domains (\textit{drawings} and \textit{towers}). Each domain is defined formally over an \textit{initial library of base program primitives} $\mathcal{L}_{base}$ , and a hierarchical \textit{generative model} to procedurally construct graphics programs for rendering object stimuli withing four nested \textit{subdomains} with varying higher-order part structure. 

This section provides additional details on these domains. The full generative models for both domains, along with the generated stimuli, are released at the repository.

\subsubsection{Drawings stimulus generation.}\hfill\\
\textit{Initial primitives $\mathcal{L}_{base}$.} The initial program primitives for the drawing domain consist of a CAD-like set of simple base shape primitives and matrix operations for transforming and combining these shapes into a final drawing:
\begin{itemize}
\setlength{\itemsep}{1pt}
\setlength{\parskip}{0pt}
\setlength{\parsep}{0pt}
    \item \texttt{line}: a unit-length horizontal line shape.
    \item \texttt{circle}: a unit-radius circle shape.
    \item \texttt{square}: a unit length square shape.
    \item \texttt{scaled\_rect(w, h)}: a rectangle parameterized by width and height.
    
    \item \texttt{graphics\_matrix(scale, theta, x, y)}: produces a transformation matrix parameterized by scale, rotation angle, and x and y translation angle.
    \item \texttt{apply\_transform(shape, matrix)}: applies a transformation matrix to a shape.
    \item \texttt{repeat(shape, n, matrix)}: Applies the same transformation successively n times to a given shape and returns all of the n shapes.
    \item \texttt{connect(shape, shape)}: Connects two shapes into a single complex shape.
    \item Mathematical operations (eg. \texttt{plus}, \texttt{minus}, \texttt{sin}) over floating constants.
    
\end{itemize}\hfill\\
\textit{Drawings generative model.} Drawings stimuli are defined over the base primitives as programs which combine subdomain-specific parts according to varied program templates, parameterized by the number and size of each part. We generate stimuli by sampling randomized parameterizations over these templates to produce the 250 stimuli sampled for each subdomain. These subdomain are described below using semantic terms for their hierarchial part structure, but these terms simply correspond to templated graphics programs parameterized recursively by size and number of subparts.

We define the following four drawings subdomains, described with a high-level overview of their underlying part structure:
\begin{itemize}
\setlength{\itemsep}{1pt}
\setlength{\parskip}{0pt}
\setlength{\parsep}{0pt}

\item \texttt{nuts and bolts}: combines parts for an \textit{outer shape} (eg. a hexagonal nut) of varying size, and \textit{inner perforation} of varying size (eg. a circular hole), and 1 or more \textit{ring perforations} of varying size (eg. a ring of circular holes.)
\item \texttt{vehicles}: combines parts for n \textit{wheels}, \textit{vehicle bases} of varying sizes and templated types, and varying numbers of parameterized \textit{antenna} or \textit{windows}.

\item \texttt{gadgets}: combines parts for n \textit{dials} or \textit{buttons}, templated \textit{gadget bases} of varying sizes, and varying numbers of \textit{antenna}.

\item  \texttt{furniture}: combines parts for n \textit{knobs}, \textit{drawers}, templated \textit{furniture bases} of varying sizes and furniture \textit{feet}.
\end{itemize}

\subsubsection{Towers stimulus generation.}
\textit{Initial primitives $\mathcal{L}_{base}$.} The initial program primitives for the structures domain build on the towers planning domain from \cite{ellis2020dreamcoder}, which consists of simple primitives for picking and placing colored blocks:

\begin{itemize}
\setlength{\itemsep}{1pt}
\setlength{\parskip}{0pt}
\setlength{\parsep}{0pt}
    \item \texttt{vertical\_red}: a unit-length vertical red block.
    \item \texttt{horizontal\_blue}: a unit-length vertical blue block.
    \item \texttt{left(n, canvas, block)}: moves a simulated cursor left n steps on a canvas and places a block.
     \item \texttt{right(n, canvas, block)}: moves a simulated cursor right n steps on a canvas and places a block.
\end{itemize}\hfill\\

\textit{Structures generative model.} Structures stimuli are defined over the base primitives using groups of low-level part abstractions: \texttt{tiles}, \texttt{arches}, and \texttt{house-parts} contained fixed arrangements of blocks; \texttt{rows(width)} and \texttt{pillars(height)} were parameterized functions.
Different subsets of these were hierarchically combined in different ways to create each subdomain.
Each subdomain was parameterized by numerical parameters (e.g. number of arches), as well as by part type parameters (e.g. wall tile).

\texttt{Bridges} contained up to 2 external arches, and up to 5 internal arches of a different type. Each low-level \texttt{arch} abstraction contained two red pillars, which could all be extended up to a maximum height. Each bridge could support a mid-level \texttt{viaduct} abstraction consisting of multiple rows. We also defined a \texttt{suspension(suspension-type)} function that placed red blocks directly above the pillars, of a \texttt{uniform} height, or that \texttt{decreased} or \texttt{increased} in height towards the center of the bridge.

\texttt{Cities} contained two skyscrapers placed a random distance apart. Each skyscraper was defined by a wall \texttt{tile}, which was  \texttt{stacked(height)} vertically and optionally \texttt{mirrored}, and topped with a \texttt{row} or \texttt{pyramid} \texttt{roof}. 

\texttt{Houses} were each topped with a \texttt{pyramid} \texttt{roof}, defined by the width of the house. The width of the house was determined by the width of the first \texttt{floor}, which could contained any permutation of \texttt{{window, bricks, door}}. Up to two additional floors contained permutations of \texttt{{window, bricks}}.

\texttt{Castles} were defined by a central mirrored \texttt{wall} of \texttt{tiles}, and two flanking \texttt{stacks} of \texttt{tiles}, each parameterized by a height (with \texttt{stack height < wall height}), and the wall additionally parameterized by width. The wall and stacks were topped with the same type of \texttt{roof}, providing there was space. Each roof could be a \texttt{pyramid} or \texttt{dome}- similar to a pyramid with an additional shorter row beneath.

We exhaustively enumerated items from each subdomain (with a small range for each parameter), rejecting any tower that extended beyond a 20x20 grid.

\section{S2. Part II Supplemental Details} \label{sec:s1_part_ii}
This section describes additional technical details for the library identification model used in \textbf{Part II} of the main paper. Code for both the defined hypothesis spaces and alignment model is also released at the provided repository.

\subsection{Defining candidate program libraries}
All stimulus from each subdomain are generated from the shared set of base program primitives $\mathcal{L}_0$ described in \textbf{S1}. However, as described in the main paper, our goal is to define candidate \textit{program libraries} $\mathcal{L}_i$ that augment the initial base primitives with \textit{additional program subroutines} which encapsulate higher-level part structure specific to a given subdomain (such as the \textit{wheel} components shared across the vehicles stimuli, or the \textit{external arch} components shared across the bridge stimuli. In particular, we aim to construct \textit{cumulatively defined libraries} such that each $\mathcal{L}_i$ contains all of the program concepts in library  $\mathcal{L}_{i-1}$, along with new program subroutines at a higher level of complexity.

While we could define an enormous set of possible candidate libraries containing all possible program subroutines across our stimulus subdomains, we consider a restricted set of representative libraries constructed from the natural hierarchical structure used to generate the underlying stimuli, and which are designed to span the maximal range of candidate part concepts on our subdomains (from the common base primitives in $\mathcal{L}_0$ to extremely complex subroutines that correspond to entire categorical classes of stimuli on each subdomain.) 

In particular, we construct a given library $\mathcal{L}_i$ to contain new primitives corresponding roughly to the 'next highest outer loop of parametric variation' over the part components in $\mathcal{L}_{i-1}$. On the nuts and bolts domain, for instance, $L_0$ contains the original \texttt{unit line} and \textit{unit circle} shape primitives, $L_1$ contains additional \texttt{polygons} defined by parameterizing rotation over the \texttt{unit lines}, $L_2$ contains additional \textbf{rings of shapes}  (including rings of polygons) defined by parameterizing rotation over all of the shapes in $L_1$, and $L_3$ contains categorical stimulus concepts (like a \texttt{outer shape with a ring of shape perforations} parameterized by the particular shape and size of its components.

This procedure is designed to maximize discriminability of the candidate libraries while still corresponding to the underlying recursive program structure of each subdomain in a principled way. However, in future work we hope to look to \textit{automated} program library construction procedures which can be used to generate a much larger space of candidate libraries, like those in \cite{ellis2020dreamcoder, wang2021learning, wong2021leveraging}.

\subsection{Library-to-vocabulary alignment model}
This provides further implementational details on the model used to produce the \textit{library-to-vocabulary alignment} metric shown in \textbf{Figure 4} of the main paper.

This metric measures the \textit{mean token log-likelihood} (which correlates negatively with perplexity) for each subdomain with respect to the set of \textit{descriptions} for each object stimulus, \textit{programs} that generate that object stimulus, and a given \textit{library} in which those programs can be represented. 

More specifically, each library $\mathcal{L}_i$ is comprised of $\rho_0^i, \rho_1^i... \in \mathcal{L}_i$. The graphics programs for each object stimulus are initially represented as a program $\pi_{base}$, but they can be rewritten automatically with respect to any given library $\mathcal{L}_i$  into a semantically equivalent program $\pi_{i}$ that maximally compresses the original program to use the encapsulated subroutines in the new library. We call the \textit{tokenization} of a program $\pi_{i}$ its left-order tree traversal (omitting variables), such that the tokenization of a program is an ordered sequence of program components $[\rho_{\pi} ...]$.

The goal of our metric is to estimate the average alignment probabilities between \textit{program components} in tokenized programs for each object stimulus (written in a given library) and the \textit{words} in the tokenized \textit{what} descriptions for that object stimulus. We use the IBM Model 1 translation model \cite{brown1993mathematics}, which takes as input a corpus of paired (program token, description token) sequences for all object stimuli in a given subdomain, and estimates corpus-wide type-type translation probabilities $P(w | \rho)$ for each word type $w$ across all descriptions, and each program type $\rho$ across all programs written under a given library. This model also jointly estimates \textit{alignments} between the individual program and word tokens in a (description, program) pair -- it finds the MAP alignments  $\alpha(w, \rho)$ such that $\Pi_{\alpha} P(w | \rho)$ is maximized for all aligned words and program components.

We report a \textit{cross-validated} alignment metric over each subdomain using batches of n=5 heldout-stimuli: we randomly order all (description, program) pairs, fit the IBM model to estimate $P(w | \rho)$ based on all but the heldout stimuli, and then report our metric with respect to the MAP alignments on the heldout (description, program) pairs. More specifically, we n each of these heldout (description, program) pairs, the mean token log-likelihood corresponds to the mean log $P(w | \rho)$ where the mean is taken over the MAP alignments $\alpha(w, \rho)$.

\bibliographystyle{apacite}

\setlength{\bibleftmargin}{.125in}
\setlength{\bibindent}{-\bibleftmargin}

\bibliography{cogsci_2022_supplemental}